\documentclass{sig-alternate}
\usepackage{wrapfig}
\usepackage{float}
\usepackage{url}
\begin{document}

%

\title{Hands-free Evolution of 3D-printable Objects via Eye Tracking
}
%
%
%
%
%

\numberofauthors{4} 
%
\author{
%
%
Nick Cheney$^\star$, Jeff Clune$^\dagger$, Jason Yosinski$^\star$, Hod Lipson$^\star$\\
nac93@cornell.edu, jclune@uwyo.edu, yosinski@cs.cornell.edu, hod.lipson@cornell.edu\\\\
$^\star$Creative Machines Lab, Cornell University, Ithaca, New York\\
$^\dagger$Evolving AI Lab, University of Wyoming, Laramie, Wyoming
}
\maketitle

\begin{abstract}
Interactive evolution has shown the potential to create amazing and complex forms in both 2-D and 3-D settings.  However, the algorithm is slow and users quickly become fatigued.  We propose that the use of eye tracking for interactive evolution systems will both reduce user fatigue and improve evolutionary success.  We describe a systematic method for testing the hypothesis that eye tracking driven interactive evolution will be a more successful and easier-to-use design method than traditional interactive evolution methods driven by mouse clicks.  We provide preliminary results that support the possibility of this proposal, and lay out future work to investigate these advantages in extensive clinical trials.  
\end{abstract}

\section{Introduction}

People are generally good critics, but are often poor at describing exactly what they want, especially for things they have never seen before.  While it is difficult to precisely describe something in technical terms, people usually find it much easier to look at a set of options and declare which ones they prefer. This phenomenon is captured by the phrase ``I'll know it when I see it'' and may be due to a lack of technical knowledge to explain a conceptualized idea or the inability to imagine something never previously encountered. Additionally, some ideas for designs seem preferable before they are viewed, while others seem undesirable, but look surprisingly good once instantiated.

Interactive evolution uses this idea to drive design, either of solutions to a particular problem \cite{sims1992interactive} or of open-ended creation where the only objective is aesthetic appeal.  The algorithm presents human users with potential solutions, and allows them to show a preference for things they like and discourage things they don't like.  It then uses the information provided from the user's feedback to create novel designs similar to previously preferred solutions, iteratively finding designs more and more preferential to the user.  For a full introduction to Evolutionary Algorithms, please refer to Goldberg and Holland 1988 \cite{goldberg1988genetic} or Goldberg 1989 \cite{goldberg1989genetic}.

\begin{figure}[t]
	\centering
		\includegraphics[width=3.3in]{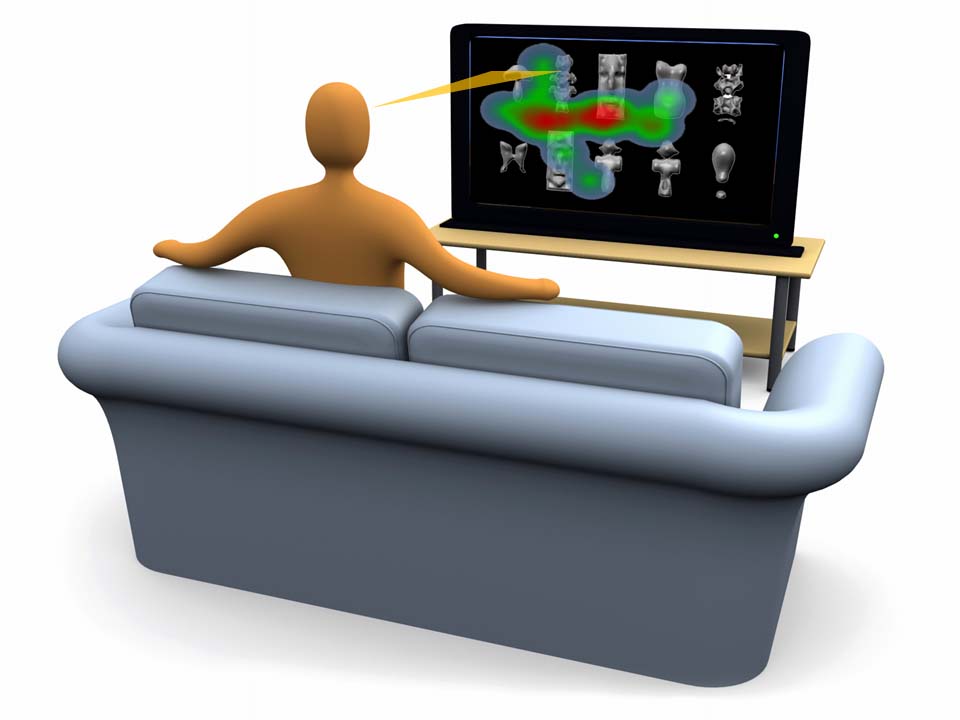}
		\caption{The appeal of interactive evolution through eye tracking becomes clear as we consider the reduction of user fatigue, implicit optimization advantages, passive nature, and increasing availability of eye tracking capable devices and distributed 3D printing technology for fabrication of these custom designs.}
		\label{onCouch}
\end{figure}

As one may imagine, interactive evolution is slow and the user becomes fatigued, which can lead to insufficient amounts of time and effort dedicated towards design.  There have been numerous efforts to relieve this fatigue by predicting user preferences and offloading some of this interactivity to a machine proxy or substitute for the human user, usually though interweaving human and computer evaluations \cite{parmee2000multiobjective, machwe2006introducing, kamalian2005reduced, gu2006capturing, hornby2012accelerating}.  However, few studies have focused on the human-computer interface itself in order to allow a greater number of trials or less overall fatigue, as each interaction is made to be less taxing than one performed with a traditional interface.  Holmes and Zanker attempted to produce attention driven evolution with an eye tracker \cite{holmes2008eye}.  However, this work suffered from limited subject data, a tightly confined design space (in which only a single number was evolved that represented the ratio of the side-lengths of a rectangle), and no comparisons of eye tracking driven evolution to evolution driven by more traditional mouse or keyboard interaction interfaces.  Pallez et al. also note the promise and importance of such a system, but lack the implementation of an eye tracker to demonstrate it \cite{pallez2007eye}.  

Advances in eye tracking based interactive evolution may be driven by the interaction between the following two observations. First, Joseph et al. found that unique stimuli --- in their setup, unique rotational orientations --- were likely to cause involuntary shifts in visual attention \cite{joseph1996involuntary}.  Second, Lehman and Stanley showed that a preference towards novel solutions may be a better driver of evolution than fitness (target-goal) based approaches, which often converge to local-optima in complex problems \cite{lehman2008exploiting}.  These two complementary properties suggest that simply displaying potential designs and measuring attention via an eye tracker may involuntarily draw attention to the most novel or unique designs shown to a user and thus provide a powerful driving force for interactive evolution.  

One of the greatest potential advantages of eye tracking is that it gathers user-feedback on all the displayed objects, not just the one or two they would end up selecting in traditional click-based selection. The user paints all of the objects with preference information via the amount of time they gaze at each object, providing much more data to the interactive evolution algorithm per generation, which should reduce fatigue and improve performance. 

Furthermore, multi-objective evolution towards both target-driven and novelty-driven solutions has been shown to produce solutions which have both high diversity as well as high performance towards the given target-driven objective, exemplified though niche exploitation \cite{lehman2011evolving}.  Thus it is possible that interactive evolution with eye tracking in target-directed design will also undergo some degree of this involuntary attention towards novel designs, further improving evolution through this implicit property of attention.  

In addition to its optimization advantages, eye tracking in itself may be an attractive interface for interactive design, as it can also enable participation from new populations of users, such as those with physical disabilities, or those using devices which traditionally do not employ interactivity (e.g. televisions), but could easily take advantage of passive or involuntary interactions.  Increasingly, consumer devices such as computers and cell phones now include the capability to incorporate eye tracking, suggesting the possibility for including passive, preference-driven, customized design as part of everyday technological interactions.

Coupled with recent innovations in 3-D printing, this method of automating design via an effortless interface will greatly increase the use the general public has for in-home 3-D printers \cite{malone2007fab}.

\section{BACKGROUND}

The ability of interactive evolution based on CPPN-NEAT \cite{stanley2007compositional} to produce ascetically pleasing, complex forms has been demonstrated both in 2-D with Picbreeder.com \cite{secretan2008picbreeder} and in 3-D with EndlessForms.com \cite{clune2011evolving}.

Lohse noted that consumers were likely to spend 54\% more time viewing Yellow Pages advertisements of business they end up choosing than those from business they do not choose \cite{lohse1997consumer}.  This suggests that using visual attention as a proxy for preference in our eye tracking setup is likely to bias selection towards designs preferred by the user.  Pieters and Warlop further supported this idea by finding that respondents tended to fixate longer on brands of products that they eventually chose, compared to alternative choices \cite{pieters1999visual}.

Rayner et al. found that consumers were likely to spend more time looking at types of ads to which they were instructed to pay attention \cite{rayner2001integrating}.  This suggests that attempts at target-driven evolution are justified to use visual attention as a proxy for intentional preference towards certain designs.  

Preliminary studies led by some of this paper's authors also tested the use of a brain-computer interface---an Emotiv EPOCH Neuro-headset---to drive evolution based on what a user thinks about the objects they see. The idea was to augment eye tracking information with the emotional reaction to the object being looked at. The study concluded that the headset and its included software is not sophisticated enough to report reliable, timely emotional-response data. Even in extreme cases, where a subject strongly liked or disliked an object, the noise in the system swamped the signal and such preferences could not be inferred via thoughts alone. Additionally, the precision and accuracy of the consumer-grade headset was not sufficient to allow users to explicitly select objects one by one.  Attempts to have the user select objects via thought control failed because of the effort, delay, and inaccuracy of the device. However, we expect that augmenting, or even solely driving, interactive evolution with information from brain-computer interfaces could be an important improvement in interactive evolution with future, improved brain-computer interfaces that are faster and more accurate. 

\begin{figure}[t]
	\centering
		\includegraphics[height=1.8in]{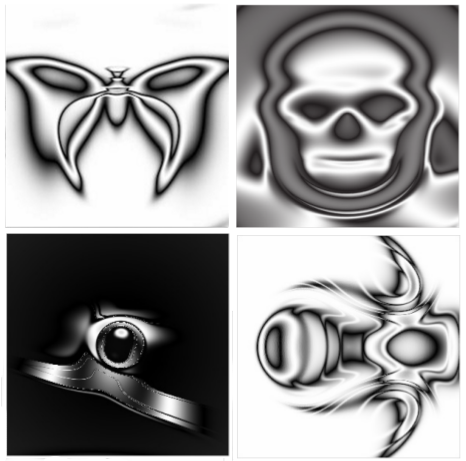}
		\hspace*{5pt}
		\includegraphics[height=1.8in]{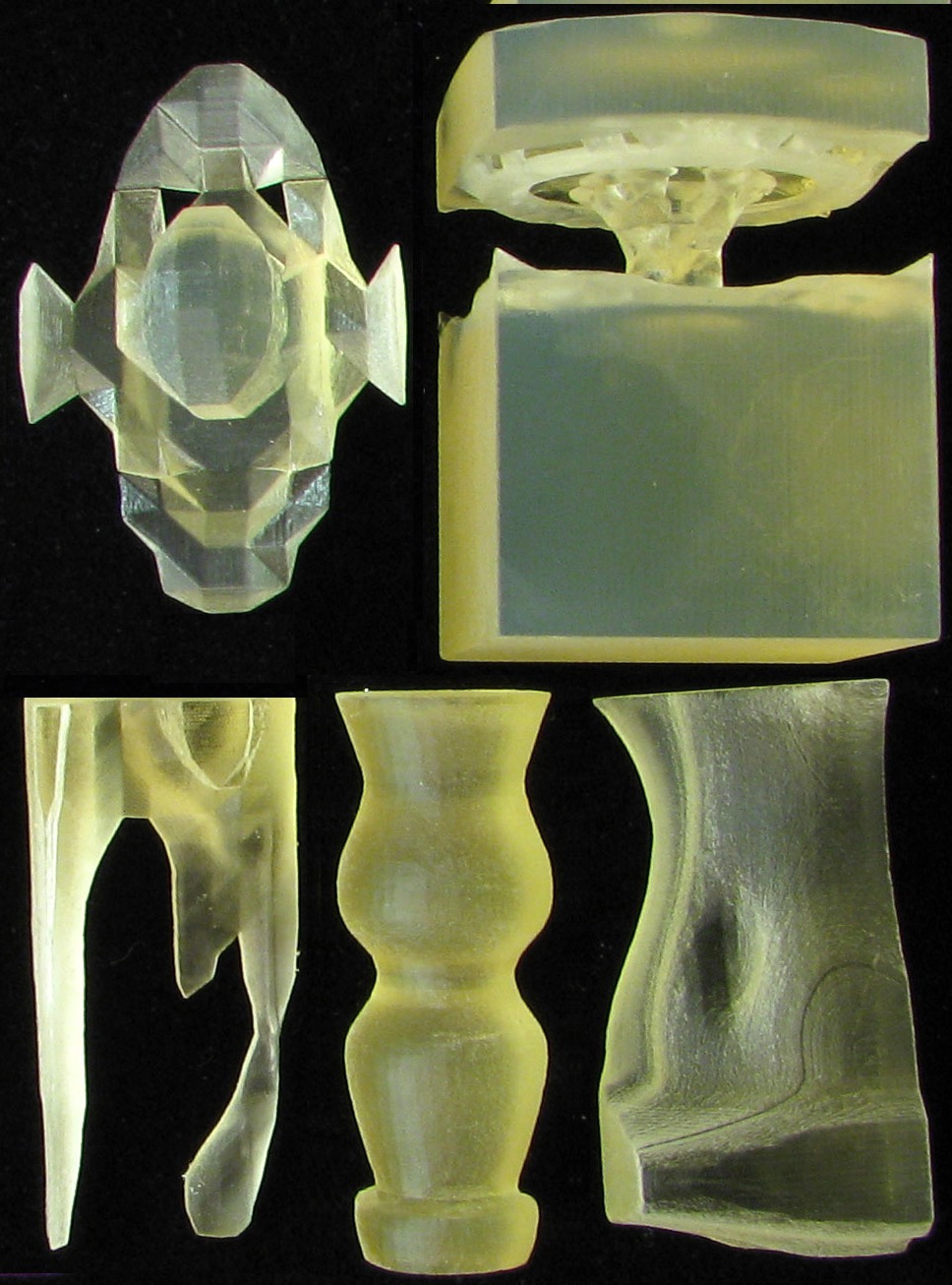}
		\caption{(left) Examples of high resolution, complex, natural-looking images evolved with CPPN-NEAT that contain symmetry, repetition, and interesting variation \cite{secretan2008picbreeder}. (right) Examples of CPPN-encoded 3-D shapes with these same properties, also demonstrating the ability to fabricate such designs via a 3D printer~\cite{clune2011evolving}.}
		\label{picbreederEndlessforms}
	\label{cppnImagesAndObjects}
\end{figure}

\section{METHODS}

\subsection{EYE TRACKING}

Subjects are placed approximately 27 inches in front of a 20-inch computer monitor set at eye level, with a Mirametrix S1 Eye tracker placed directly below the monitor.  Eye tracker calibration is performed by the Mirametrix calibration software, in which the user focuses on a blue dot as it moves to nine different locations on the screen.  

The eye tracker operates by shining infrared light into the eye to create reflections that cause the pupil to appear as a bright, well-defined disc in the eye tracker camera. The corneal reflection is also generated by the infrared light, appearing as a small, but sharp, glint outside of the pupil. The point being looked at is then triangulated from the corneal reflection and the pupil center \cite{poole2006eye}.

If, during the eye tracking portions of the trials, the pupils left the capture range of the eye tracker camera or the point-of-regard appeared off-screen, the system would pause, only to resume upon the return of a valid, on-screen signal.  

In preliminary tests with subjects and the authors, the object the system determined the user was looking at was illuminated. All users remarked that the system was highly accurate and nearly always illuminated the object they were looking at. 

Eye tracking trials occurred in two distinct stages: directed design and free-form design.

\subsection{STAGE 1: DIRECTED DESIGN}

Subjects are given a specific target objects (e.g. small, red cones) and asked to try to evolve such a shape. Targets are randomly selected by choosing one attribute at random (e.g. red) per attribute type (e.g. color). The attribute types and their possible values are \emph{size} \{small, large\}, \emph{color} \{red, blue, green\}, and \emph{shape} \{cone, oval, rectangle\}.  To avoid oversampling data from specific target shapes, subjects are never asked to evolve the same exact target, but their targets may share properties (e.g. if a given subject is asked to produce small red cones, he or she will not be asked to produce small red cones again, but could be asked to produce large, red cones).  

Subjects are then shown a $3\times5$ array of three-dimensional objects, each rotating around its vertical axis.  They are instructed to pay attention to the objects that most closely resemble the target object they are attempting to create.

Once the subject feels he or she has seen an object that meets the criteria of the target object (or feels that the target object is unreachable and gives up), he or she is instructed to press a button on the keyboard to signify this event.  At this time, a screen-shot of the objects on the screen is recorded and the trial is terminated.  Subjects also have the option to record the objects on screen any any time without terminating the trial by pressing a different button the keyboard.  For each subject, this process is repeated three times with unique target objects.

After each attempt to create a target a brief questionnaire asks the subject about their feelings towards completion of their target and their reason for terminating the trial.  At the termination of the stage (after the third trial questionnaire), the subject is presented with a short survey.  Here, the subject is asked questions about their enjoyment, ease, and feelings of successful task completion.  The survey is filled out on a computer and answers to each question are either on the five-point Likert Scale or consist of a brief, free-form response.
  
\subsection{STAGE 2: FREE-FORM DESIGN}

After the completion of Stage 1, the subjects are told they will have 20 minutes for a free-form design session. Their only instruction is simply to direct attention to the objects that they find most interesting, and explore what they can create.  

Like in Stage 1, the subjects may terminate the session and start again at any point, but there is no specific criteria for them to do so in this case.  New trials of free-form design restart upon termination of the previous trial, until such an event happens following the 20 minute mark, upon which time the stage is completed.

Screen captures behaved identically to Stage 1. Screen shots could be taken by the user at any time, and were always taken when when a subject restarted (terminated) a trial. 

Again, after each trial within the stage, a brief questionnaire asks the subject about their feelings towards completion of their open-ended goal to design interesting objects and their reason for terminating the previous trial.  Upon the completion of Stage 2, the subjects were also presented a similar survey to that presented after Stage 1, to gauge their feelings regarding enjoyment, ease-of-use, and success in creating objects.

\subsection{MEASUREMENTS}

During each trial, the object arrays shown to the user are captured.  Additionally their responses to the questions at the end of each trial and stage are kept on record.  This data is integrated with automatically collected data, including a unique numeric identifier for each subject, the amount of time and number of generations spent on each trial, and number of resets in the 20 minute free-form section.

\begin{figure}[t]
	\centering
		\includegraphics[width=3.3in]{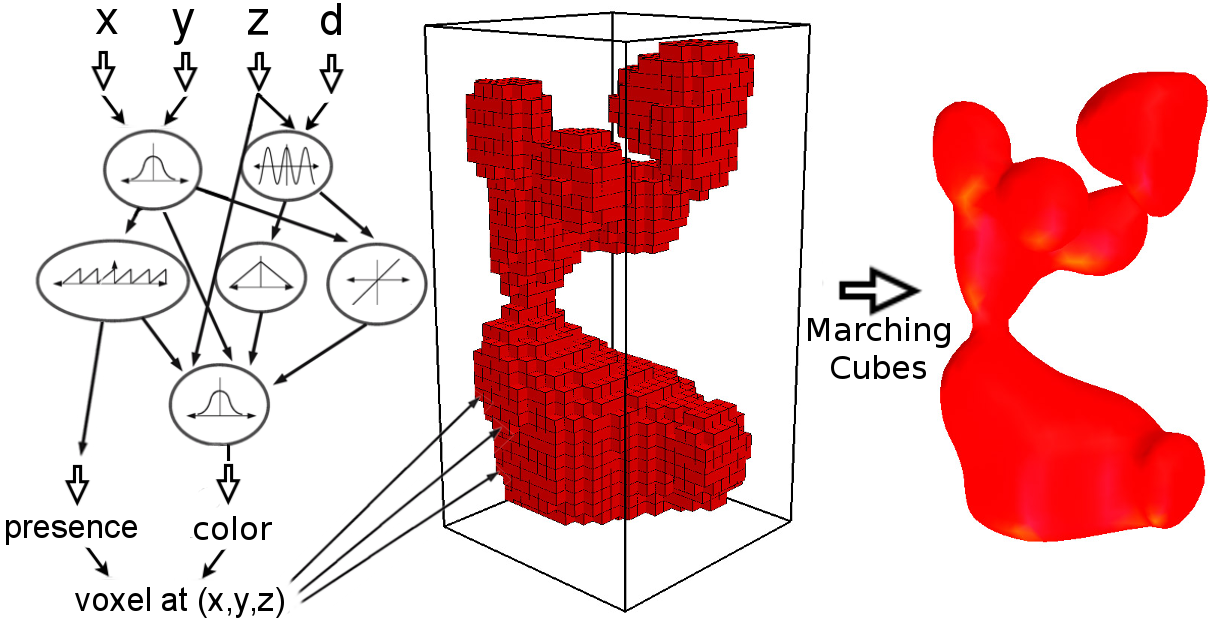}
		\caption{CPPN-NEAT iteratively queries each voxel within a specified bounding area and produces output values as a function of the coordinates of that voxel. These outputs determine the shape and color of an object.  The voxel shape is then smoothed with a Marching Cubes algorithm to produce the final object. Although the CPPN-NEAT network is queried at some finite resolution, it actually specifies a mathematical representation of the shape and thus, critically for high quality 3-D printing, it can be queried with arbitrarily high resolution.}
		\label{cppnDepiction}
\end{figure}

\vspace*{1pt}
\subsection{OBJECT CREATION: CPPN-NEAT}

Consistent with \cite{secretan2008picbreeder} and \cite{clune2011evolving}, we employ CCPN-NEAT to encode and evolve the object designs.  CPPN-NEAT has been repeatedly described in detail~\cite{stanley2007compositional, clune2011performance, clune2011evolving, gauci2007generating}, so we only briefly summarize it here. A compositional pattern-producing network (CPPN) is a way to encode designs in the same way nature encodes its designs (e.g. overlapping chemical gradients during the embryonic development of jaguars, hawks, or dolphins). A CPPN is similar to a neural network, but its nodes contain multiple math functions (in this paper: sine, sigmoid, Gaussian, and linear). CPPNs evolve according to the NEAT algorithm~\cite{stanley2007compositional}. A CPPN produces geometric output patterns that are built up from the functions of these nodes. Because the nodes have regular mathematical functions, the output patterns tend to be regular (e.g. a Gaussian function can create symmetry and a sine function can create repetition). These patterns specify phenotypic attributes as a function of their geometric location. In this paper, each voxel has an $x$, $y$, and $z$ coordinate that is input into the network, along with the voxel's distance from center ($d$). An output of the network, queried at each geometric-coordinate location, specifies whether any material is present at a given location. The 3 remaining output nodes are queried once (at the center point) and specify the RGB values that comprise the object's color~(Fig.~\ref{cppnDepiction}). By producing a single CPPN representing the functional structure of a design, and iteratively querying it for each voxel, we can produce the entire structure of the object at any resolution.\\

\subsection{EVOLUTIONARY PARAMETERS}

The Evolutionary Algorithm \cite{goldberg1988genetic, goldberg1989genetic} employed here is NEAT (NeuroEvolution of Augmenting Topologies) \cite{stanley2002evolving}.  A population size of 15 is used, such that the entire population is displayed to the user at each generation in the on-screen $3\times5$ array.  There was no explicit cap on the number of generations, as the trials were terminated by user command.  NEAT speciation was not employed in this setup.

In the experimental (eye tracking) setup, at each refresh loop of the algorithm and on-screen images, if the eye tracker records the user's point-of-regard within the $3\times5$ array cell corresponding with that individual, that individual would gain the clock time since the last refresh loop of the algorithm (this value is typically a small fraction of a second).  This process lasts until one individual organism accumulates one second (1000 milliseconds) of time it was looked at.  At this point, the generation ceases, and each individual is assigned the fitness equal to the time it was looked at during that generation (in milliseconds).  Thus the top individual at each generation would have a fitness of 1000, while all other individuals have a fitness between 999 and 1 (the minimum baseline fitness), depending on the time the user spent looking at each of the designs during the given generation. \\

\begin{figure}[t]
	\centering
		\includegraphics[width=3.3in]{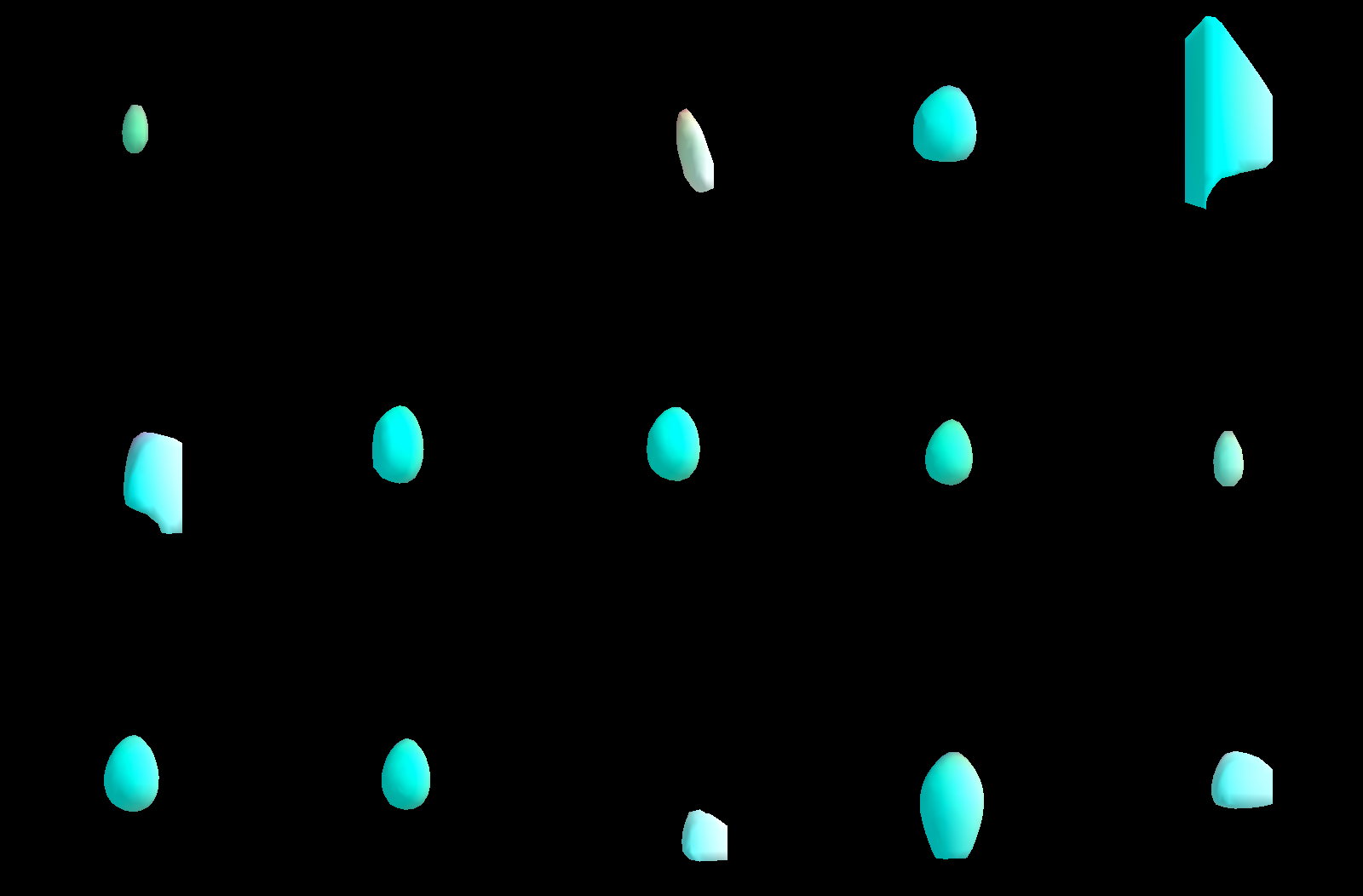}
		\caption{The results of a directed design period in which the target design was a small blue oval.  The subject produced this array of designs after 39 generations.}
		\label{smallBlueOval}
\end{figure}

\begin{figure}[h]
	\centering
		\includegraphics[width=3.3in]{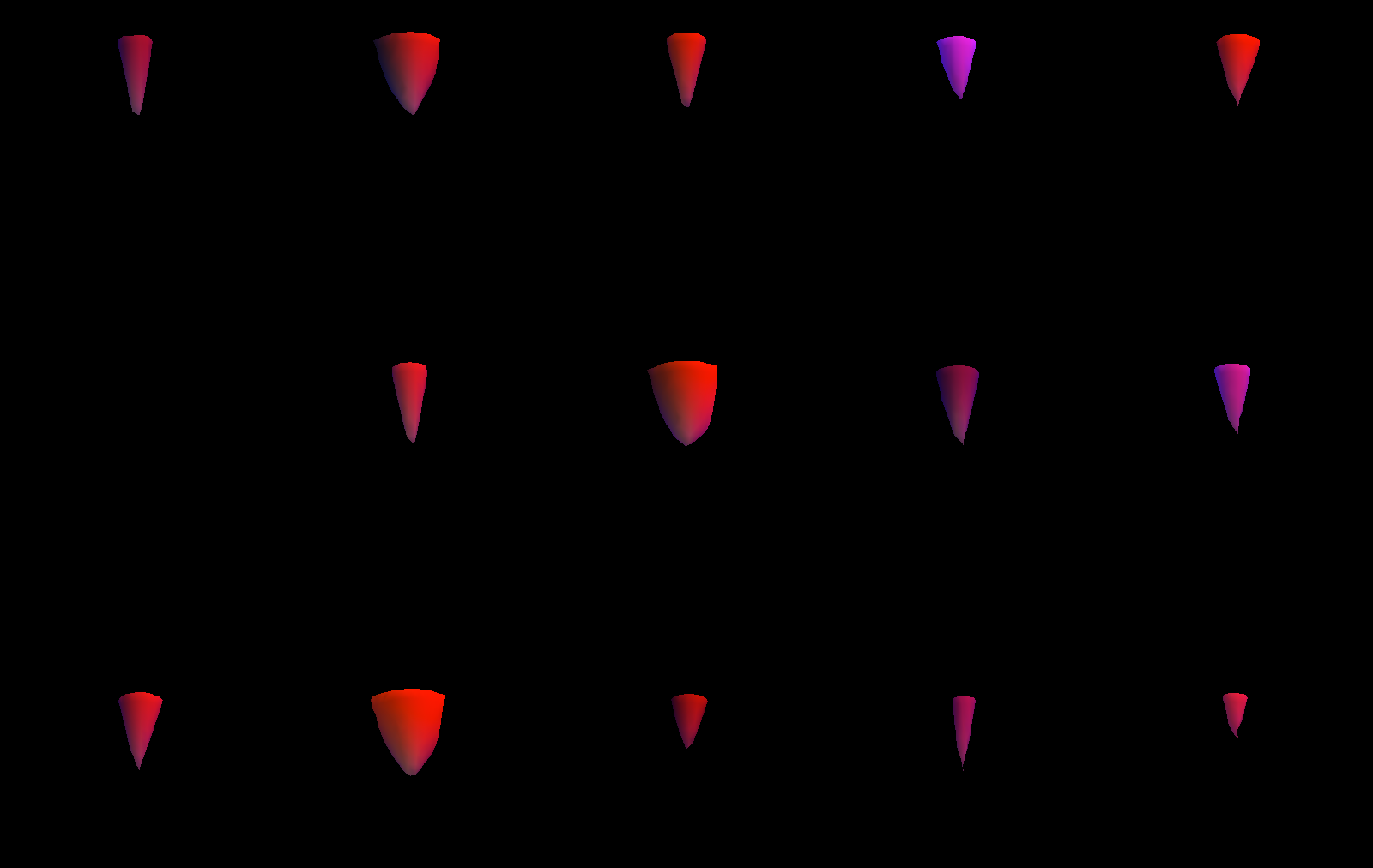}
		\caption{The results of a directed design period in which the target design was a small red cone.  The subject produced this array of designs after just 3 generations.}
		\label{smallRedCone}
\end{figure}

\subsection{CONTROL}

In order to control for the effects of the eye tracking system on interactive evolution, a control setup is also created which mirrors the eye tracker setup with a few exceptions.  All individual designs are manually selected by the user via a computer mouse.  To reflect the ability of the visual system to explore multiple designs, users are allowed to right-click on multiple designs to explicitly indicate a preference for more than one design before left-clicking to submit the selections.  
 Selected individuals are given a standard fitness value (1000), while non-selected organisms are left with the baseline fitness value of 1. There is no automatic progression from generation to generation without a user-provided mouse click in this setup.  The instructions and questionnaires are slightly different to accommodate the use of mouse instead of eye tracking.  

Subjects are exposed to both the experimental (eye tracking) setup and the control (mouse-clicking) setup.  The order in which the control or experimental conditions are presented are counterbalanced. The target-based stage always preceded the free-form phase for both the experimental and control treatment.

\section{PRELIMINARY RESULTS}

Preliminary results show the ability of a subject to create the target shape with just their eyes.  While a complete data set from clinical trials has yet to be collected, in these preliminary experiments a majority of target-directed trials resulted in subjects who felt that they had successfully reached their target goals using the eye tracking system. Fig.~\ref{smallBlueOval} shows design towards small, blue ovals.  The subject considered the target to be successfully reached after 39 generations.  Fig.~\ref{smallRedCone} shows design towards small, red cones.  The subject considered the target to be successfully reached after only 3 generations in this case.  

\begin{figure}[t]
	\centering
		\includegraphics[width=3.3in]{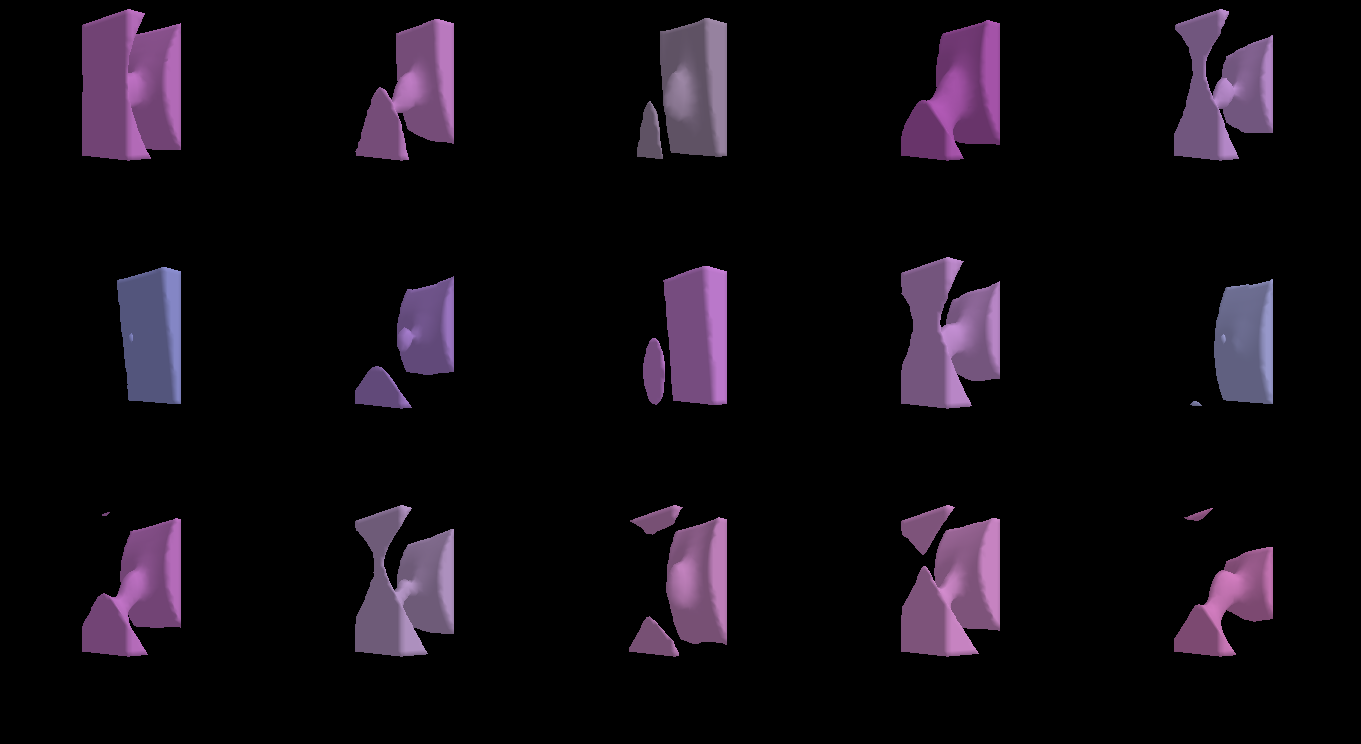}
		\caption{Example designs produced during the free-form exploration period.  With no pre-specified objective, the subject produced designs that appear similar to a computer monitor.}
		\label{monitor}
\end{figure}

\begin{figure}[t]
	\centering
		\includegraphics[width=3.3in]{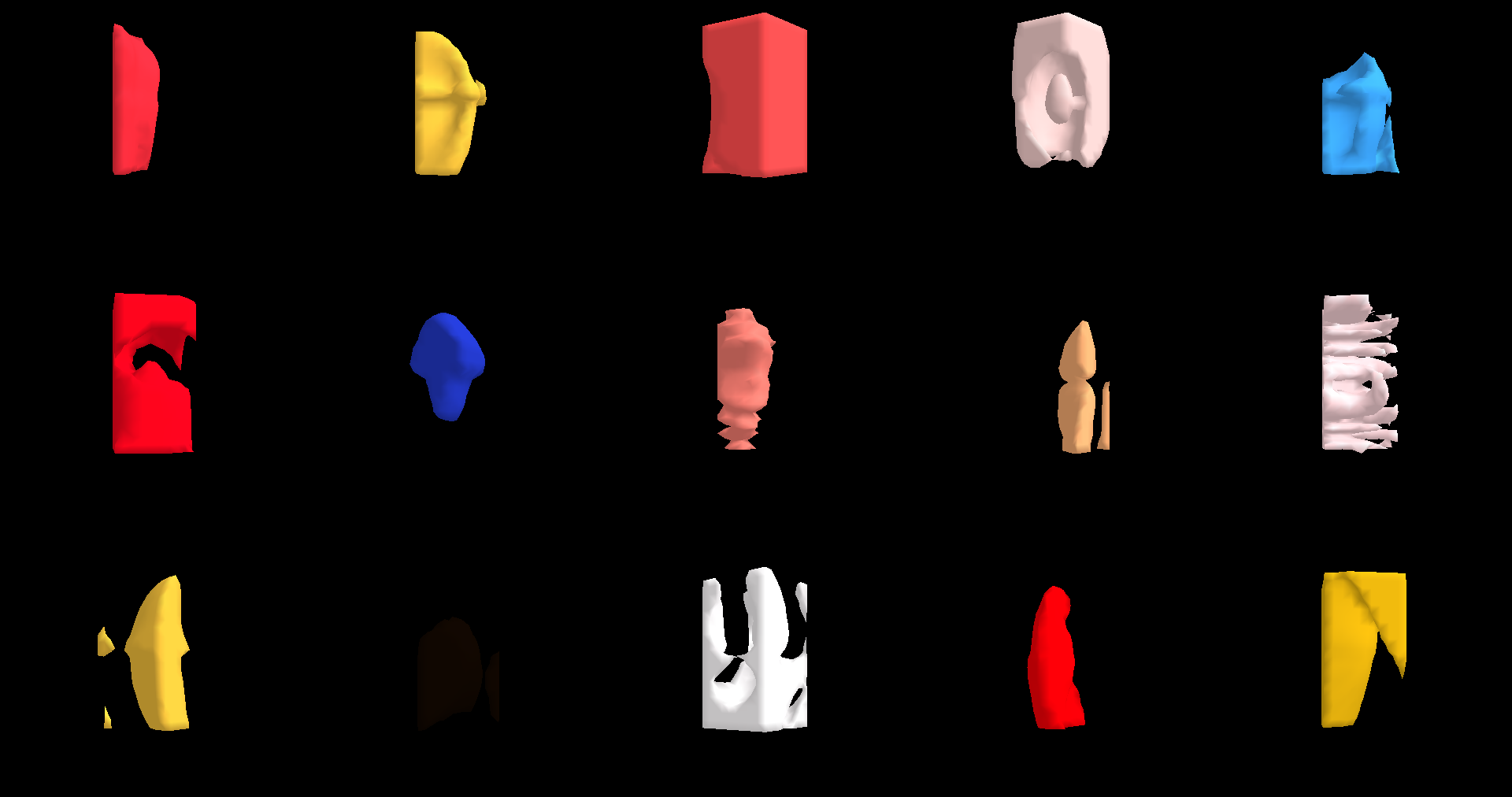}
		\caption{Never-before-seen shapes and ideas can be created through this open-ended process.  Here is an example of unique designs produced by a subject during the free-form exploration period.}
		\label{freeForm}
\end{figure}

Free-form design also shows promising preliminary results.  Fig.~\ref{monitor} demonstrates a set of designs produced by a subject which are reminiscent of a computer monitor.  While no target object was specified for this trial, an idea for a common and physically useful object design was created from the subject simply paying greater attention to interesting objects on the screen.  While this open-ended process can lead to the rediscovery of known shapes, it can also lead to completely novel ones.  Fig.~\ref{freeForm} shows a set of designs that do no converge towards any particular shape, but simply explore ideas the user found interesting.

\section{FUTURE WORK}

An in-depth study of the technologies introduced in this paper is in progress. While the goal of this work is to provide a proof-of-concept, along with anecdotal data that show working examples, the in-depth study will provide substantial quantitative measurements of what subjects can accomplish with this technology.  Specifically, the follow-up paper will include many trials, analyze the resulting complexity of evolved designs, report the clock-time and number of generations spent creating each design, and summarize the self-reported feelings of the subjects about using the system. This information will be provided for both the eye tracking experimental treatment as well as the mouse-click-driven control treatment.  It will also explore both target-directed and free-form interactive evolutionary design in both setups.  This study, which is currently ongoing, will be the first systematic demonstration of the advantages of eye tracking based interactive design over traditional user interfaces for open-ended design.

The future potential of eye tracking for interactive design is enormous - especially when one considers its potential for commercial use.  The ability to customize design without the need to formally describe it will open many doors for distributed design, which will couple well with the increases in distributed fabrication as 3D printers become commonplace appliances.  

However a large number of questions still need to be addressed.  For example, user attitude and mindset are also important, as users classified as maximizers behave and focus attention differently than those considered to be satisficers in consumer choice theory \cite{chowdhury2009time}.  The extent to which this aspect of individual variation will affect the selection process of interactive evolution is still unclear, as is the extent to which user conditioning can modulate these effects in the setting of interactive evolution.  

Additionally, while visual attention on a larger scale is incorporated into this model, saccades (the quick eye movement around a scene or object) and their effect on transitions between individual designs or eye movements within a single $3\times5$ grid cell are not accounted for in this model. That is because the Mirametrix technology ignores them to assess the point the user is consciously looking at. Nor is it clear to what extent the grid resolution and the accuracy of the eye tracker affect the capability of this system to capture the full set of useful information that is available.  These saccadic eye movements have been shown to interact with visual attention in certain settings \cite{hoffman1995role}, so accounting for them in an attention driven system such as this may provide new insights.  

\section{CONCLUSION}

In this work we describe a new way to allow people to design objects. This technology allows hands-free design by passively gathering user feedback via eye tracking. We demonstrate that the technology does accurately infer which object the user is looking at and can use that information to direct successful design sessions. The preliminary data reported suggests that users can successful design target objects and can produce interesting, novel shapes without touching a keyboard or mouse. Future, clinical studies will provide a rigorous investigation of the advantages and disadvantages of eye tracking over traditional interactive evolution interfaces.\\\

\section{ACKNOWLEDGMENTS}

This work was funded by NSF CDI Grant ECCS 0941561NSF and Postdoctoral Research Fellowship in Biology to Jeff Clune (DBI-1003220).  

%
\bibliographystyle{abbrv}
\small{\bibliography{Hands_Free_Evolution_via_Eye_Tracking--Cheney_Clune_Yosinski_Lipson}}  
%

\end{document}